\documentclass[conference,compsoc]{IEEEtran}
\usepackage{cite}
\usepackage{hyperref}
\hypersetup{
    colorlinks=true,
    linkcolor=black,
    anchorcolor=black,      
    urlcolor=black,
    citecolor=black,
}
\usepackage{graphicx}
\usepackage{epsfig}
\usepackage{subfig}
\graphicspath{{images/}}
\DeclareGraphicsExtensions{.pdf, .jpg, .jpeg, .png, .eps}
\usepackage{amsmath, bm, amssymb}
\usepackage{makecell}
\usepackage{tabularx}
\usepackage{multicol}
\usepackage{multirow}
\usepackage{diagbox}
\usepackage[export]{adjustbox}
\begin{document}
\title{Mul-GAD: a semi-supervised graph anomaly detection framework \\via aggregating multi-view information}
\author{\IEEEauthorblockN{Zhiyuan Liu}
\IEEEauthorblockA{Scholar of Cyberspace Security\\Hainan University\\
Haikou, Hainan 570228--0898\\
Email: fyhvyhj@gmail.com}
\and
\IEEEauthorblockN{Chunjie Cao}
\IEEEauthorblockA{Scholar of Cyberspace Security\\Hainan University\\
Haikou, Hainan 570228--0898\\
Email: caochunjie@hainanu.edu.cn}
\and
% \IEEEauthorblockN{James Kirk\\ and Montgomery Scott}
\IEEEauthorblockN{Jingzhang Sun}
\IEEEauthorblockA{Scholar of Cyberspace Security\\Hainan University\\
Haikou, Hainan 570228--0898\\
Email: jingzhangsun@outlook.com}}
\maketitle
% 异常检测背景
% 当前challenge
% 提出的方法以及简要介绍做法，解决challenge
% 在数据集上的实验结果
\begin{abstract}
Anomaly detection is defined as discovering patterns that do not conform to the expected behavior. Previously, anomaly detection was mostly conducted using traditional shallow learning techniques, but with little improvement. As the emergence of graph neural networks (GNN), graph anomaly detection has been greatly developed. However, recent studies have shown that GNN-based methods encounter challenge, in that no graph anomaly detection algorithm can perform generalization on most datasets. To bridge the tap, we propose a multi-view fusion approach for graph anomaly detection (Mul-GAD). The view-level fusion captures the extent of significance between different views, while the feature-level fusion makes full use of complementary information. We theoretically and experimentally elaborate the effectiveness of the fusion strategies. For a more comprehensive conclusion, we further investigate the effect of the objective function and the number of fused views on detection performance. Exploiting these findings, our Mul-GAD is proposed equipped with fusion strategies and the well-performed objective function. Compared with other state-of-the-art detection methods, we achieve a better detection performance and generalization in most scenarios via a series of experiments conducted on Pubmed, Amazon Computer, Amazon Photo, Weibo and Books. Our code is available at https://github.com/liuyishoua/Mul-Graph-Fusion.
\end{abstract}
\IEEEpeerreviewmaketitle

\section{Introduction}
% 1. anomaly detection and graph neural network
% 2. limitation of conventional shallow method, neural network and the superior of graph neural network  
% 3. The emergency of current graph-based method. And newest exploring in the ensemble learning (one problem)
% 4. Determine the scene of model(semi-supervised learning). The solution divided into two parts: loss function designing and fusion scheme. (two solution for the problem)(3,4 togeother)
% 5. Experiment on loss function and fusion scheme show
% 6. To rationalize our attempt on fusion scheme, we formulate fusion motivation with a Venn diagram, which shows mathematical or logical connections between different groups of things.
% 7. contributions of our studies
As a long-standing and critical subject, there is a surge of attention interests on anomaly detection. Recent researches have widely explored on social security field, such as Credit fraud\cite{credit1,credit2,credit3}, Rumor detection\cite{rumor1,rumor2}, Network Intrusion\cite{network_intrusion} etc, which is directly related to civil livelihood. Industry in the security-critical field, such as medical, military and national security, even regard it as a critical obstacle. Previous efforts mostly focused on shallow learning, but with little improvement. Meanwhile, as the explosion increasing of graph-based method purposed, modeling anomaly pattern using graph-based method has gradually become the mainstream paradigm. Due to powerful modeling capabilities among local individuals, graph neural network based (GNN-based) methods not only offer a significant improvement on anomaly detection, but also gain lots of attention.

Briefly, anomaly can be defined as patterns that do not conform to expected behavior. Put differently, the patterns deviate significantly from the majority of normal ones. According to the anomalous definition, experts formalize unified anomaly detection paradigm and develop lots of functional methods. For a general taxonomy, anomaly detection method can be divided into shallow learning and GNN-based methods. Shallow learning methods mostly detect outlier from assuming the prior distribution of data or observing the difference of spatial density surrounding normal and anomaly samples. Since the obviously inductive bias and lack of the ability to capture the non-linearity relation, shallow learning methods can only learn the shallow anomaly pattern of data and are hard to work effectively in the complex cases. More specifically, the shallow methods are able to detect the frequent anomaly behavior individuals, but lose the power to detect the relatively normal or less anomaly behavior ones. Differently, since modeling the relation of each individual is one of the properties of GNN-based methods, the disguised anomalies can be detected indirectly via the information derived from their surroundings. Leveraging the superiority of nonlinear modeling of neural networks and the relational modeling of graph structure, GNN-based methods gain increasing attention.

% 现有解决方案并引出我的方法
Although GNN-based methods behave more superiority than both shallow learning, existed studies\cite{motivation1,motivation2} illustrate that no GNN-based anomaly detection method shows generalization performance on most datasets. Some attempts on multi-view learning\cite{mul_view,mul_view_community} try to bridge the gap via merging heterogeneous information or mixing with community analysis method. In contrast, our multi-view solution considers improving generalization performance under the setting of fixed information entropy, which means we optimize on the method level rather than adding additional data.

\begin{figure}[!t]
    \centering
    \includegraphics[width=0.48\textwidth, right]{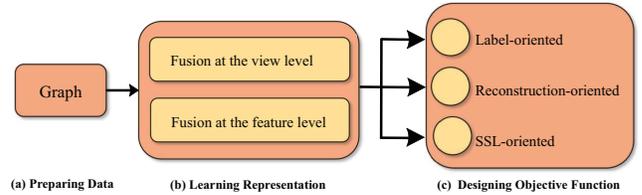}
    \caption{The pipeline of anomaly detection framework.}
    \label{fig:overview}
    \vspace{-0.5cm}
\end{figure}

% 具体方法
Graph anomaly detection pipeline can be divided into three parts: preparing data, learning representation and designing the objective function as shown in Fig. \ref{fig:overview}. Since the model is trained on the open source graph dataset, no additional graph generation operations are required. In the part of representation, we obtain multi-view representations via multiple GNN-based methods\cite{GCN,GAT,GIN,BWGNN} and transform them to a unified representation via our fusion strategies. For the objective function, label-oriented, which is often used in the semi-supervised setting, will be selected for a better generalization. After scaling the unified representation to the required size, we initialize the label-oriented objective function and update the entire networks using stochastic gradient descent. It is worth noting that our method is under the setting of semi-supervised for a more realistic scenes.

% 实验效果
As mentioned previously, our Mul-GAD model consider the optimization in two perspective, how to choose objective function and how to design fusion strategies. We summarize the objective functions of the existing graph anomaly detection and categorize them into label-oriented (semi-supervised learning), reconstruction-oriented (unsupervised learning), and ssl-oriented (self-supervised learning) according to the division of supervised learning. Experiments show that label-oriented function has a more generalized performance in most scenarios. For the fusion solution, we optimize at view and feature levels. To control the contribution of each view, we set the learnable parameters to learn the importance of each view. Due to utilize the plentiful information from different views, the model shows a better generalization. In the feature level, we utilize the feature similarity matrix to make full use of complementary information, while avoid the influence of redundant information. Experiments show that computing the feature similarity matrix plays an importance role in boosting the detection performance. The final model, which equipped with label-oriented objective function and fusion strategies, has a significant improvement on most datasets. For a more comprehensive conclusion, we further dive into the influence of the number of fused views on detection performance.

% 方法的理论解释以及论文的贡献
Adequate experimental validation is the foundation for selecting the objective function. In terms of fusion strategies, it can be explained theoretically via the Venn diagram, which is often applied to show mathematical or logical connections between different groups of things. Our contributions can be summarized as follow:
\begin{itemize}
    % 据我们所知，第一个从损失函数的视角对异常检测问题进行分析
    \item To the best of our knowledge, we are the first to analyze anomaly detection problem from the perspective of the objective functions and find that label-oriented function have a more generalized performance.
    % 提出两种有效的融合方案
    \item We provide two effective fusion strategies at the view and feature level. Both of them boost detection performance.
    % 不一样的理论角度来解释融合方案的有效性
    % \item Utilizing the properties of Venn diagram, we elaborate the effectiveness of our fusion strategies theoretically.
    % 最终提出的方法很好latex4
    \item The Mul-GAD approach outperforms the state-of-the-art not only on detection performance, but also in terms of generalization across the majority of datasets. 
\end{itemize}
% \hfill mds
% \hfill August 26, 2015
\begin{figure*}[!t]
    \centering
    \includegraphics[width=\textwidth]{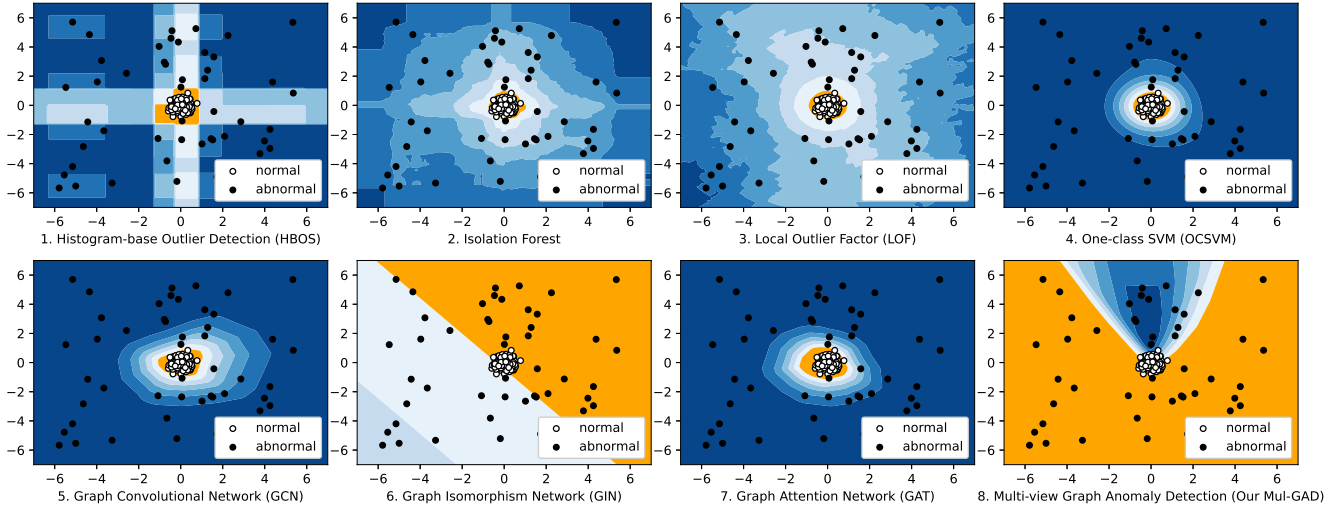}
    \caption{Decision boundary for varying algorithms. The normal and abnormal samples are generated by normal distribution $\mathcal{N}(0,0.3)$ and uniform distribution $\mathcal{U}(-7,7)$ respectively.}
    \label{fig:decision_boundary}
    \vspace{-0.4cm}
\end{figure*}
\section{Related work}
% 按两种划分进行介绍
In this section, we will introduce from the algorithm and objective function aspects. For the former, anomaly detection can be divided into shallow learning and graph neural network methods. For objective function, anomaly detection can be categorized to label-oriented, reconstruction-oriented and ssl-oriented.
\subsection{Shallow learning}
% 空间密度，统计分布，以及机器学习三个方面展开
Shallow learning, which differs from deep learning, handles anomaly problem from spatial density, statistical distribution and variants of classical machine learning methods. In general, there are fewer nodes or lower node density around the anomalies. Spatial density methods develop based on the hypothesis, a.k.a. intuition. Local outlier factor (LOF \cite{breunig2000lof}) acquires the rank of anomaly scores via computing the spatial density of each node and the lower density corresponds to a higher anomaly score. K-nearest neighbor (KNN \cite{ramaswamy2000knn}) seeks out the k closest neighbors and uses the majority class to determine the class of the current node. Constrained by the inductive bias, such methods are hard to spot the abnormal nodes masquerading as the normal ones. Statistical-based algorithms distinguish anomalies via the assumed prior distribution or computing the relevant statistical indicators. Kernel density estimation (KDE \cite{latecki2007kde}) maintains kernel function for each node. The final estimation function can be obtained via computing the mean of all kernel functions and further acquires the anomaly scores. Histogram-based outlier score (HBOS \cite{goldstein2012hbos}) conducts histogram modeling for each feature and anomaly scores can be obtained via integrating anomaly degrees for all features. Gaussian mixture model (GMM \cite{aggarwal2017gmm}), which is similar with KDE method, fits the distribution of data via multiple single Gaussian models. And the class of new coming nodes are determined by the mixed Gaussian model. Although detecting anomalous nodes from a statistical view has beneficial on the simplicity of the algorithm and the fast operation speed, its validity is highly dependent on the prior assumption for the given data. Anomaly detection tasks are naturally unbalanced binary classification tasks. Many variants of machine learning methods have been designed to adapt to the kind of imbalanced binary classification task. One-class support vector machine (OC-SVM \cite{scholkopf2001oc_svm}) derived from Support vector machine projects the raw data into an infinite-dimensional Hilbert space and distinguish anomalies via maximizing the soft margin. Isolation forest (IForest \cite{liu2008isolation_tree}) originated from decision tree utilizes binary search trees to isolate samples. However, these methods are still difficult to preform well due to the lack of nonlinear modeling capabilities.
\subsection{Graph neural network}
Graph neural network can be divided into message propagation, auto-encoder and ensemble modeling methods. Although the message propagation algorithms use the skeleton of graph neural network, such as Graph convolutional network (GCN \cite{GCN}), Graph attention network (GAT \cite{GAT}) or Graph Isomorphism Network (GIN \cite{GIN})), there is no essential difference with the traditional multi-classification models. It regards anomaly detection as an unbalanced binary classification problem and sets weights on cross entropy function to alleviate the impact of imbalance of samples. Thinking over from the frequency perspective, tang et.al purposed Beta Wavelet Graph Neural Network (BWGNN \cite{BWGNN}) which fulfilled tailored spectral filter via employing the beta distribution as wave filter. Differently, the auto-encoder methods reconstruct node attributes and topology with corresponding decoder functions, and the reconstruction errors of nodes are exploited to discover anomalous nodes on the attribute network. One of the classical representatives of the reconstruction techniques is Deep Anomaly Detection on Attributed Networks (Dominant \cite{dominant}). To capture the structure pattern effectively, anomaly detection through a dual autoencoder (AnomalyDAE \cite{dual_anomalydae}) is boosted via applying additional attention mechanisms on structure encoder to capture the significance between a node and its neighbors. However, recent studies \cite{motivation1,motivation2} have shown that both message propagating and auto-encoder methods fail to express generalization performance on most datasets. By combining the strengths of various models, the ensemble methods target to resolve the generalization problem. A Deep Multi-view Framework for Anomaly
Detection (ALARM \cite{mul_view_community}) integrates heterogeneous attribute characteristics through multiple graph encoders. Haghighi et.al \cite{mul_view_community} considers ensemble learning from the perspective of multi-task learning and employs the community-based representation technique to better capture the topology structure. Despite the great advances made by the existing ensemble methods, researches on how to fusion representation and how to design objective function remain in its adolescence. Our Mul-GAD framework was exactly purposed to bridge the generalization gap from the aspect of fusing representation and designing the objective function. As shown in Fig. \ref{fig:decision_boundary}, our fusion method has the modeling capability of GCN, GIN, GAT and is able to learn their decision boundaries.

\subsection{Objective function}
% 1. 为什么单独提出目标函数的分类
% 2. 公式确定模型前面的骨架f(X,A),
Since the design of the objective function is a crucial component of the model pipeline, it deserves a corresponding status in the modeling. However, few studies have adequately considered the design of objective function. After analyzing the existing objective functions on the anomaly detection, we discovered that the objective functions may be classified into the following three categories: label-oriented \cite{GCN,GAT,GIN,BWGNN}, reconstruction-oriented \cite{dominant,dual_anomalydae} and ssl-oriented \cite{ssl1,ssl2contrastive}, according to the division of supervised learning. Label-oriented treats the anomaly detection as a unbalanced binary classification problem and is equivalent to the weighted cross entropy. After transforming the unified representations to the required attribution and topology size, the reconstruction-oriented function can be established via computing mean square error of the reconstruction matrix. SSL-oriented function derives from contrastive learning \cite{ssl2contrastive,contrastive} and learns representation by automatically constructing similar and dissimilar instances. A detailed formulation will be carried out in next section.

\section{Methodology}
% 目前的图异常检测模型，根据这个模型，现有的研究大多专注于那个方向
% 按照我的pipeline划分，一部分是模型的表征学习，一部分损失函数的选择。
% 表征学习是方法，引入集成表征学习的概念，并进一步介绍视角和特征层面的融合
% 损失函数将三种总结的公式化。
In this section, we first formulate the graph anomaly detection and    disassemble the research framework to representation learning and objective function designing from the aspect of model pipeline. Our solutions on the two parts will be detailed below.
\subsection{Problem Definition}
We define capital letters, bold lowercase letters and lowercase letters to denote matrices, vectors and constants respectively for simplicity of comprehension, e.g. $D,\bm{d},d$. Given $\mathcal{G}\left(V,A,X\right)$ as a graph, where $V$ denotes a series of nodes $V=\left\{\bm{v_1},\bm{v_2},...,\bm{v_n}\right\}$, $n$ is the number of nodes. $X\in \mathbb{R}^{n\times d}$ indicates the attribute matrices composed by the feature vector of nodes $\left\{\bm{x_1},\bm{x_2},...,\bm{x_n}\right\}$, where d refers to the dimensional of $\bm{x}_i$. $A \in \mathbb{R}^{n\times n}$ is the adjacency matrices of $V$ where $a_{ij}$ equals to 1 if there is an edge between the i-th and j-th node, otherwise 0. Assuming that the labels of the partial nodes of graph $\mathcal{G}$ are known as $\bm{y_L}=\left\{y_{l1},y_{l2},..,y_{lp}\right\}$, $\bm{y_U}=\left\{y_{u1},y_{u2},..,y_{uq}\right\}$ represents the unknown labels that we have to deduce. $p,q$ refer the number of known and unknown labels respectively, and $p+q=n$. Therefore, $\bm{y}=\bm{y_L}\cup\bm{y_U}$ stands for the entire labels of $\mathcal{G}$. Anomaly detection framework can be formulated as follow:
\begin{equation}
\label{equation_1}
    \mathcal{F}:\mathcal{G}\left(V,A,X\right)\xrightarrow[]{}\bm{y}, \quad\bm{y}\in \mathbb{R}^n
\end{equation}

Existing approaches are in an effort to discover excellent algorithmic mappings $\mathcal{F}$ through various perspectives. As long as the algorithm is determined, the labels $\bm{y}$ can be inferred from the mapping. Following the pipeline of anomaly detection, we implement this mapping function $\mathcal{F}$ by combining two functions $\mathcal{F}_1$ Eq.\ref{equation_2}, $\mathcal{F}_2$ Eq.\ref{equation_6}, which stands for representation learning and objective function designing respectively.

\subsection{Representation learning}
The goal of representation learning is to obtain the final representation as follow:
\begin{equation}
\label{equation_2}
    \mathcal{F}_1:\mathcal{G}\left(V,A,X\right)\xrightarrow[]{}X_{f}, \quad X_{f} \in\mathbb{R}^{n\times d_f}
\end{equation}
where $d_f$ is the dimension of the final representation. In this part, we obtain the final representation by fusing multiple models. Specifically, we initially acquire the representations on each model below.
\begin{equation}
\label{equation_3}
\begin{aligned}
    &\mathcal{F}_{1-1}:\mathcal{G}\left(V,A,X\right)\xrightarrow[]{}X_{1}, \quad X_{1} \in\mathbb{R}^{n\times d_1}\\
    &\mathcal{F}_{1-2}:\mathcal{G}\left(V,A,X\right)\xrightarrow[]{}X_{2}, \quad X_{2} \in\mathbb{R}^{n\times d_2}\\
    &...\\
    &\mathcal{F}_{1-m}:\mathcal{G}\left(V,A,X\right)\xrightarrow[]{}X_{m}, \,\,\, X_{m} \in\mathbb{R}^{n\times d_m}
\end{aligned}
\end{equation}
where $\left\{\mathcal{F}_{1-1},\mathcal{F}_{1-2},...,\mathcal{F}_{1-m}\right\}$ stands for $m$ algorithms to extract characteristics, which may be GCN\cite{GCN}, GAT\cite{GAT} or BWGNN\cite{BWGNN}, etc. After that, the multiple representations would be integrated at the view and feature level.
\subsubsection{Fusion at the view}
In this subsection, the critical issue is how to fusion representations $\left(X_1,X_2,...,X_m\right)$ that differ in dimension. It can be formalized as follows:
\begin{equation}
\begin{aligned}
\label{equation_4}
    \mathcal{F}_{1-view}:\left(X_1,X_2,...,X_m\right)\xrightarrow[]{}X_{v}, \quad X_{v} \in\mathbb{R}^{n\times d_{v}}
\end{aligned}
\end{equation}
where $X_{v}$ refers to the fused representation and $d_{v}$ is its feature dimension. Specifically, we first unify them into the same dimension $\mathbb{R}^{n\times d_{unified}}$ by a fully connected network, after which we set learnable weights $\left(\alpha_1,\alpha_2,...,\alpha_m\right)$ for each representation and sum them up to acquire $X_{v}$.
\begin{figure}[t]
    \centering
    \includegraphics[width=0.48\textwidth]{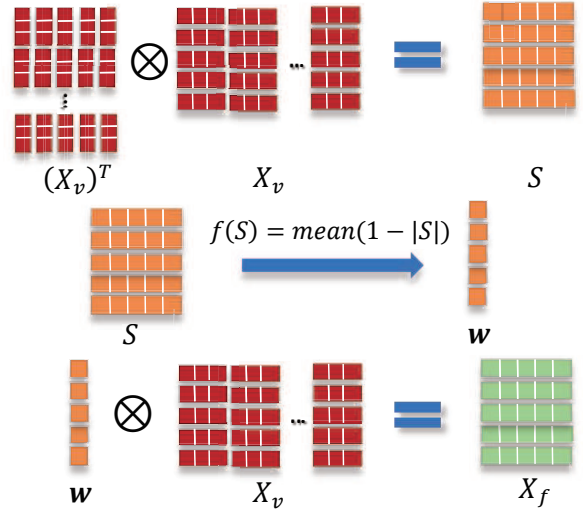}
    \caption{The fusion solution at the feature level. The first row calculates the cosine similarity among features, the second yields the importance of each feature and the third gains the final representation.}
    \label{fig:feature_method}
    \vspace{-0.4cm}
\end{figure}
\subsubsection{Fusion at the feature}
Despite the fact that we have obtained fused representations $X_{v}$ at the view level, there is still redundant information among features. The key of fusion at the feature level is to make full use of the complementary information, while avoiding the redundant information. The formula is below.
\begin{equation}
\label{equation_5}
    \mathcal{F}_{1-feat}:X_{v}\xrightarrow[]{}X_{f}, \quad X_{f} \in\mathbb{R}^{n\times d_f}
\end{equation}
where $X_{f}$ stands for the fused feature at the feature level and $d_f$ is its corresponding dimension. The combination of $\left\{ \mathcal{F}_{1-1},...,\mathcal{F}_{1-m},\mathcal{F}_{1-view},\mathcal{F}_{1-feat}\right\}$ is equivalent to the mapping function $\mathcal{F}_{1}$ and $X_{f}$ is exactly the final representation mentioned in Eq.\ref{equation_2}. In this part, we model the target of making full use of complementary information as well as avoiding redundant information by calculating feature similarity. Assuming that $S\in \mathbb{R}^{d_v\times d_v}$ and $\bm{w} \in \mathbb{R}^{d_v}$ are the feature similarity matrix and the weight vector respectively, we model the mapping $\mathcal{F}_{1-feat}$ procedure as follows:
\begin{equation}
\begin{aligned}
\label{equation_6}
    &s_{ij}=\frac{\left(X_v\right)_i\cdot \left(X_v\right)_j}{\|\left(X_v\right)_i\|\cdot \|\left(X_v\right)_j\|}, \quad 1\leq i,j \leq d_v\\
    &w_i = \frac{1}{n}\cdot\sum_{j=1}^{n}\left(1-\left|s_{ij}\right|\right), \;\quad -1\leq s_{ij}\leq 1\\
    &\left(X_f\right)_i = w_i\cdot \left(X_v\right)_i, \quad\quad\qquad 0\leq w_{i}\leq 1
\end{aligned}
\end{equation}
where $(X_v)_i \in \mathbb{R}^n$ refers to the $i$-th column of the matrix $X_v$ and $s_{ij}$ denotes the cosine similarity between the $i$-th and $j$-th columns. $S$ and $\bm{w}$ are composed of $\left\{s_{11},...,s_{ij},...,s_{d_vd_v}\right\}$,$\left\{w_1,...,w_i,...,w_{d_v}\right\}$ respectively. The reason we subtract $\left|s_{ij}\right|$ from $1$ is to construct an intuition, which the larger the value of $1-\left|s_{ij}\right|$ is, the more irrelevant the $i$-th and $j$-th columns of $X_v$ are. Therefore, they share richer complementary information and it is more preferable to assign a larger weight for the column features, which $1-\left|s_{ij}\right|$ can exactly accomplish. To obtain the redundancy strength of a single feature over the whole feature, we acquire $w_i$ by calculating the mean of $S$ in row or column. The larger $w_i$ means that the $i$-th column features have richer complementary information and less redundant information, which can be multiplied with $\left(X_v\right)_i$ to get the final representation $\left(X_f\right)_i$. Fortunately, the above E.q \ref{equation_6} can be reformulated by matrix, which enables acceleration by GPU. For the comprehension, you can refer to Fig. \ref{fig:feature_method}.

\subsection{Objective function designing}
The target of objective function designing is to deduce the unknown labels as follow:
\begin{equation}
\label{equation_7}
    \mathcal{F}_2:X_{f}\xrightarrow[]{}\bm{y}, \quad\bm{y}\in \mathbb{R}^n
\end{equation}
where $\bm{y}=\bm{y}_L\cup\bm{y}_U$ denotes the whole labels of $\mathcal{G}$ including $\bm{y}_L$ labels already known and $\bm{y}_U$ labels to be deduced. We can implement E.q.\ref{equation_7} by the following three objective functions: label-oriented, reconstruction-oriented and ssl-oriented. It is worth noting that since the objective functions are classified from a supervised learning perspective, they cannot be used by stacking simultaneously. After acquiring experimental evidence, we will choose the well-performed objective functions as part of our approach.
\subsubsection{Label-oriented}
Due to the semi-supervised learning, $\bm{y}_L$ will be used for monitoring model training.
% \begin{equation}
\begin{align}
\begin{gathered}
\label{equation_8}
    X_f\xrightarrow[]{f(\cdot)}\bm{p},\quad \bm{p}\in\mathbb{R}^{n}\\
    \min_{f(\cdot)}-\frac{1}{|\bm{y_L}|}\sum_{i=1}^{|\bm{y_L}|}\left(\lambda\cdot\bm{y}_i\cdot\log\bm{p}_i+(1-\bm{y}_i)\cdot\log(1-\bm{p}_i)\right)
\end{gathered}
\end{align}
% \end{equation}
where $\bm{p}$ denotes the predicted anomaly scores of nodes and $\bm{p}_i$ refers to the $i$-th element of $\bm{p}$. By compressing $X_f$ from $d_f$ to 1 dimension through multilayer perceptron, we are able to obtain $\bm{p}$. Afterwards, $f(\cdot)$ will be updated by minimizing the cross-entropy of $\bm{y}_L$ with respect to the prediction score $\bm{p}$. Notably, we evade the impacts of sample imbalance by adjusting the balance factor $\lambda$. The unknown label $(\bm{y}_U)_i$ can be inferred as anomaly if $\bm{p}_i$ is larger than 0.5, otherwise normal.
\subsubsection{Reconstruction-oriented}
Since it belongs to the scope of unsupervised learning, we do not adopt any known labels.
\begin{equation}
\begin{aligned}
\label{equation_9}
    &X_f\xrightarrow[]{f(\cdot)}\left\{
    \begin{aligned}
        X_{attr} & , &X_{attr} \in \mathbb{R}^{n\times d}\\
        A_{stru} & , &A_{stru} \in \mathbb{R}^{n\times n}
    \end{aligned}
    \right.\\\\
    &\min_{f(\cdot)}\|X_{attr}-X\|_2+\lambda\cdot\|A_{stru}-A\|_2
\end{aligned}
\end{equation}
where $X_{attr}$, $A_{stru}$ are equivalent to $X$, $A$ after construction and $\lambda$ is the trade-off between the attribute and structure matrices. We optimize the multilayer perceptron $f(\cdot)$ by minimizing the reconstruction errors of the attribute and structure matrices. For a specific node $\bm{v_i}$, we obtain its anomaly score via calculating $\|(X_{attr})_i-X_i\|_2+\lambda\cdot\|(A_{stru})_i-A_i\|_2$. Depending on the ranking of the anomaly scores, the nodes can be classified as normal or abnormal, thus obtaining the whole labels $\bm{y}$.
\subsubsection{SSL-oriented}
Self-supervised learning does not expose known labels $\bm{y}_L$, however constructs positive and negative instances to learn representations. Specifically, we apply Deep Graph Infomax (DGI\cite{ssl2contrastive}) to initialize the SSL-oriented objective function.
\begin{equation}
\begin{gathered}
\label{equation_10}
    X_f\xrightarrow[]{f(\cdot)}\left\{
    \begin{aligned}
        H & , &H \in \mathbb{R}^{n\times d_h}\\
        \hat{H} & , &\hat{H} \in \mathbb{R}^{n\times d_h}
    \end{aligned}
    \right.\\
    \bm{s}=\sigma\left(\frac{1}{n}\sum_{i=1}^{n}\bm{h}_i\right)\\
    \min_{f(\cdot)}-\frac{1}{2n}\sum_{i=1}^{n}\left(\log\mathcal{D}(\bm{h}_i,\bm{s})+\log\left(1-\mathcal{D}(\bm{\hat{h}}_i,\bm{s})\right)\right)
% \end{aligned}
\end{gathered}
\end{equation}
where $H$, $\hat{H}$ consist of $\left\{\bm{h}_1,\bm{h}_2,...,\bm{h}_n\right\}$,$\{\bm{\hat{h}}_1,\bm{\hat{h}}_2,...,\bm{\hat{h}}_n\}$ respectively and $\mathcal{D}$ is a discriminator which produces the affinity score for each local-global pair. $\sigma$ indicates the sigmoid activation function. With the mapping function $f(\cdot)$, $X_f$ is projected to two identical algebraic spaces and yields $H$, $\hat{H}$. Obtaining from the pooling operation, $\bm{s}$ can be regarded as the global representation of $H$. Thereby, the constructed positive instance $H$ has a higher affinity with s than the negative instance $\hat{H}$. By modeling the intuition, it is possible to accomplish self-supervised learning in the no-label case and we are able to obtain a well-performed representation $H$ by updating $f(\cdot)$. To adapt the detection task, we connect a two-dimensional linear classifier after $H$. Afterward, anomaly scores can be gained by fine-tuning $f(\cdot)$ as well as training the linear model.
\begin{figure*}[!t]
    \centering
    \includegraphics[width=\textwidth]{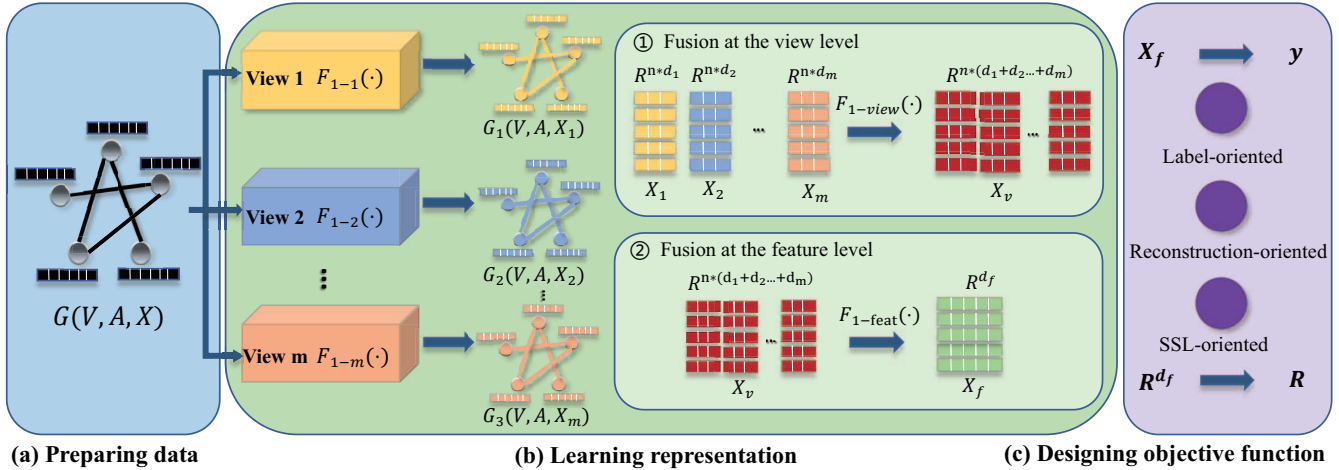}
    \caption{The overview of framework of anomaly detection model Mul-GAD. It is divided into three main modules: (a) preparing data, (b) learning representation and (c) designing objective function.
    After processing (a) and (b), we constrict the representations to the specified size to adapt the objective function. Finally, we update the model parameters by minimizing the objective function.}
    \label{fig:method}
    \vspace{-0.4cm}
\end{figure*}
\subsection{Mul-GAD detection method}
As shown in Fig. \ref{fig:method}, the pipeline of our Mul-GAD method can be divided into three parts: (a) preparing data, (b) learning representation and (c) designing objective function. In this paper, no additional graph data preparing operation is required due to the use of open source datasets. However, data preparation is particularly critical in the specific business scenarios, since the quality of data determines the upper bound of the algorithm. For the learning representation (b), We first obtain multi-view representations through multiple GNN-based methods (GCN\cite{GCN}, GAT\cite{GAT}, BWGNN\cite{BWGNN} or others), followed with fusion solutions at the view and feature level. View-level fusion is conducted to measure the contribution of each method, while feature-level fusion leverages complementary information and avoid the impact of redundant information. After the fusion representations, we need to choose the objective function (c) from three commonly used graph anomaly detection objective functions (label-oriented, reconstruction-oriented and ssl-oriented). Experiments show that the label-oriented objective function performs well and has a more generalization performance on most datasets. Therefore, the Mul-GAD method selects the label-oriented objective function, which is typically used in the semi-supervised learning setting. The specific formula can be obtained in the previous subsection.
\begin{figure}[tp]
    \centering
        \includegraphics[width=0.45\textwidth]{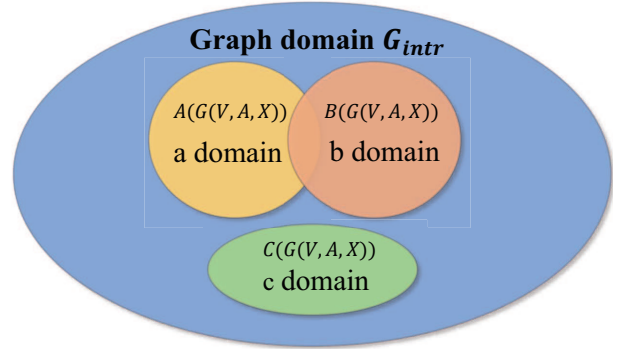}
    \caption{Schematic diagram to explain the motivation of our fusion solution.}
    \label{fig:venn_diagram}
    \vspace{-0.4cm}
\end{figure}
\subsection{Venn diagram explanation}
In general, it is believed that there exists upper bound of the algorithm for a given data. Hence, the model has a performance bottleneck and We formulate the description as:
\begin{align}
    \label{equation_tradition}
    \forall f(\cdot),\;P\left(f\left(\mathcal{G}(V,A,X)\right)\right)<\alpha,\;0<\alpha\leq1
\end{align}
where $f(\cdot)$ indicates the detection algorithm and $P$ refers to the detection accuracy. Even if we iterate through the entire algorithm space, the detection accuracy cannot exceed the upper bound $\alpha$.
Thinking differently, we suppose that there exists “\textit{intrinsic representation}” of $\mathcal{G}$, which is able to reach the upper bound $\alpha$ and can be separated by the simplest linear model. We formulate as follows:
\begin{align}
    \label{equation_different}
    P\left(f_{linear}\left(\mathcal{G}_{intr}\right)\right)=\alpha
\end{align}
Thus, the goal turns to seek for the “\textit{intrinsic representation}” of $\mathcal{G}$, i.e. $\mathcal{G}_{intr}$. Venn diagram as shown in Fig. \ref{fig:venn_diagram}, which is often applied to show mathematical or logical connections between different groups of things, is adopted to explain the motivation of our fusion solutions. The “\textit{intrinsic representation}” can be regarded as the blue area of $\mathcal{G}_{intr}$. Using the $\mathcal{A}(\cdot)$, $\mathcal{B}(\cdot)$, $\mathcal{C}(\cdot)$ algorithm on the original graph $\mathcal{G}(V,A,X)$ yields $a$, $b$, $c$ subdomain. $a$, $b$ subdomains overlap due to the similarity property of the algorithm (the overlapping parts denote redundant information), while $c$, being independent of $a$ and $b$, has sufficient complementary information. These sub-domains may be contextual, structured, frequency information or whatever. Therefore, an intuitive idea comes out whether we can gather more complementary information by means of multi-task learning, while designing modules to avoid the influence of redundant information. Our fusion solutions are derived from the intuitive, striving to approximate the dataset domain $G_{intr}$.
\begin{table*}[!t]
\renewcommand{\arraystretch}{1.3}
\caption{Statistics of the datasets. Min\_ratio, syn\_ratio denote the anomalous ratio generated by the minimal class and synthetic methods respectively, while org\_ratio means the organic ones. The number of anomalies is described in the brackets.}
\label{table.graph_data}
\centering
\begin{tabularx}{0.85\textwidth}{cccccccc}
\Xhline{2\arrayrulewidth}
\textbf{Graph} & \textbf{Nodes} & \textbf{Edges} & \textbf{Features} & \textbf{Classes} & \textbf{Min\_ratio} & \textbf{Syn\_ratio} & \textbf{Org\_ratio}\\
\hline
Pubmed\cite{pubmed} & 19,717 & 88,648 & 500 & 3 & 20.81\% (4,103) & 10\% (1,972) & -\\
% \hline
Amazon Computer\cite{amazon} & 13,752 & 491,722 & 767 & 10 & 2.12\% (292)& 10\% (1,375)& -\\
% \hline
Amazon Photo\cite{amazon} & 7,650 & 238,162 & 745 & 8 & 4.33\% (331)& 10\% (765)& -\\
% \hline
Books\cite{books} & 1,418 & 3,695 & 21 & 2 &- & - & 1.97\% (28)\\
% \hline
Weibo\cite{weibo} & 8,405 & 407,963 & 400 & 2 & - & - & 10.33\% (868)\\
\Xhline{2\arrayrulewidth}
\end{tabularx}
\end{table*}
\section{Experiments and Results}
In this section, we purpose four valuable research questions.
\begin{itemize}
\item \textbf{RQ1}: How do the different objective functions perform?
\item \textbf{RQ2}: What is the impact of two fusion solutions on detection performance?
\item \textbf{RQ3}: Whether the more views are fused, the higher the detection rate improvement?
\item \textbf{RQ4}: Can the final proposed method solve the problem that existing methods are hard to generalize on most datasets?
\end{itemize} 
To address these issues, we design a series of experiments. We initial with a basic experimental setup, following with experiments to explore the impact of the objective function and the two fusion methods on performance. To obtain rigorous conclusions, ablation study is performed to discover the effects between the two fusion methods. Moreover, the impact of the number of fusion methods is studied. As a result, our method will be compared with the state-of-the-art methods.
\subsection{Experimental setting}
% 数据做表格，表格中给引用
\subsubsection{Datasets}We used five popular real-world attributed graph datasets, which are public datasets widely used in previous studie\cite{motivation1, amazon_use_in_other_paper, min_class3}. The anomalous tags in Weibo and Books are already labeled, while Pubmed, Amazon Computer and Amazon Photo generate anomalies by human construction since there is no ground truth of anomalies in these datasets. The mainstream way of constructing is minimal class and synthetic methods. Detail statistics are listed in Table. \ref{table.graph_data}.
\begin{itemize}
    \item \textbf{Pubmed}\cite{pubmed} dataset consists of 19717 scientific publications related to diabetes in the PubMed database with 44338 links.
    \item \textbf{Amazon Computer and Photo}\cite{amazon} are segments of the Amazon co-purchase graph \cite{amazon_sub}, where nodes represent goods, edges indicate that two goods are frequently bought together.
    \item \textbf{Weibo}\cite{weibo} is a hashtag graph of user posts from a Twitter-like platform. It has 8,405 users and 61,964 subject tags. The weight of the user-hashtag is the number of specific hashtags posted by the user.
    \item \textbf{Books}\cite{books} is a second graph from the Amazon network, in which we use tags provided by Amazon users.
\end{itemize}
\textit{Minimal class anomalies} are made by treating the smallest number of classes in the original data as the anomalous class and the rest as normal, following with \cite{min_class1,min_class2,min_class3} works. \textit{Synthetic anomalies} can be obtained by generating structure and attribution anomalies, following with \cite{syn_attr_anomalies,syn_structure_anomalies}. \textit{Organic anomalies} have been manually marked, which requires no additional manipulation. For a fair comparison, we divide the training, validation and test sets as 4:3:3 on all datasets. The anomalous ratio for synthetic anomalies is set to 0.1, while the other two do not require to be set since the anomalous ratio is originally owned. The reason we use minimal class and synthetic anomalies when we have organic anomalies is to demonstrate that our methods have outstanding generalization whether on simulated or real data. In this paper, we obtain the synthetic anomalies via A Python Library for Graph Outlier Detection (Pygod\cite{pygod}).
\subsubsection{Baseline}We compare with two categories of baselines, shallow learning and GNN-based method. According to the division of shallow learning, we employ the LOF\cite{breunig2000lof} as a representative of spatial density method, the HBOS\cite{goldstein2012hbos} to represent the statistical distribution method and IForest\cite{liu2008isolation_tree}, OC-SVM\cite{scholkopf2001oc_svm} to stand for the variant of the classical machine learning method. To avoid duplicate wheel building, we adopt Python Outlier Detection (PyOD\cite{pyod}) to implement these algorithms and use the default parameters. For the GNN-based approach, we use the commonly used backbone GCN\cite{GCN}, GIN\cite{GIN} and GAT\cite{GAT} to extract the feature information of the graph. In addition, considering the excellent performance of BWGNN\cite{BWGNN} on the most datasets, it will also be applied as a baseline. It is worth noting that all the GNN-based methods establish label-oriented objective functions to optimize the network parameters.
\subsubsection{Parameter Setting}For the shallow learning baseline, we use the default hyper-parameters in PyOD library\cite{pyod}. In the GNN-based setting, we apply the two-layer networks, following \cite{GCN,BWGNN,two_layer}. Learning rate is set to 5e-3 and the dimension of the hiddle layer is set to 64. A Method for Stochastic Optimization (Adam\cite{adam}) is adopted as the optimizer, where the weight decay is set to 5e-2. For a fair comparison, we employ early stop strategy, which stops the training if the loss of the validation set does not increase in 20 epochs. This allows the models to converge to an optimal level. To enhance the computational efficiency of graph modeling, we use the PyTorch Geometric\cite{pyg} to construct the graph network.
% 实验使用的数据集，包括生成异常和原生的异常数据集。使用这两种的目的
% 模型骨架的网络层数，follow谁的研究使用两层即可。
% 优化器使用adam（引用）。for a more realistic case.
% 为了公平的比较，学习率，训练轮次都设置相同。使用5次后的均值更加严谨，并收集训练期间验证集最高的性能，代表模型的潜能。
% 介绍baseline，并确定训练设定的参数

% 常用的有accuracy, recall，precision，f1以及AUC-ROC。因为...原因，评价指标统一成AUC-ROC。Receiver operating characteristic/Standard deviation
\subsubsection{Evaluation}
In general, accuracy, recall, precision, f1 score and the area under the curve (AUC) are often applied on multi-classification tasks. Since anomaly detection is an unbalanced binary classification task, accuracy does not precisely describe the detection performance of the algorithm. And some anomaly detection algorithms can only obtain the ranking of the anomaly scores instead of the anomaly probability of the current sample, which leads to the inability to compute recall, precision and f1 score. Hence, we adopt AUC as an evaluation metric for anomaly detection, which is obtained via computing the area under the receiver operating characteristics. To acquire stable results, we take the mean value over five runs in all experiments. And the standard deviation (STD) is used to describe the stability of the results. In addition, we obtain the best result of the validation set during the training process to describe the potential of the model.
\begin{table}[!t]
\renewcommand{\arraystretch}{1.6}
\caption{The detecting results of the different objective function AUC/STD (\%) over five runs among various backbones, on two types of anomalies for Pubmed.}
\label{table.Objective function choose}
\centering
\setlength{\tabcolsep}{1mm}{
\begin{tabularx}{0.48\textwidth}{c|cccc}
\Xhline{2\arrayrulewidth}
\diagbox[]{Type.}{Loss.}&Backbone&Rec-oriented&SSL-oriented&Label-oriented\\
\hline
\multirow{4}{2em}{Min}&GIN&89.2$\pm$0.2&92.2$\pm$0.0&\textbf{92.3$\pm$0.0}\\
&GCN&88.5$\pm$0.4&89.8$\pm$0.2&\textbf{91.7$\pm$0.1}\\
&GAT&80.3$\pm$15.1&90.0$\pm$0.3&\textbf{92.2$\pm$0.0}\\
&BWGNN&53.5$\pm$1.4&86.4$\pm$15.9&\textbf{95.9$\pm$0.0}\\
\hline
\multirow{4}{2em}{Syn}&GIN&74.2$\pm$0.4&78.3$\pm$0.9&\textbf{80.1$\pm$0.0}\\
&GCN&68.9$\pm$4.9&73.5$\pm$0.2&\textbf{76.7$\pm$0.8}\\
&GAT&57.4$\pm$8.2&75.0$\pm$2.7&\textbf{86.5$\pm$0.3}\\
&BWGNN&69.2$\pm$1.5&83.6$\pm$0.3&\textbf{84.7$\pm$0.8}\\
\Xhline{2\arrayrulewidth}
\end{tabularx}}
\vspace{-0.5cm}
\end{table}
\subsection{Influence of The Objective Function}
To explore the impact of different objective functions on the detection performance \textbf{RQ1}, we compare various GNN-based methods as shown in Table. \ref{table.Objective function choose}. Rec-oriented in the table denotes reconstruction-oriented. The results show that the label-oriented objective function is about 3 to 4 percentage points higher than the others. Moreover, its detection performance is more stable due to the lower standard deviation. Hence, in the case of GNN-based skeletons, we do not recommend the usage of reconstruction-oriented and ssl-oriented. Instead, label-oriented is advised to adopt in an unknown situation. Our final Mul-GAD method will adopt label-oriented as our objective function.
\begin{figure}[!t]
    \centering
    \includegraphics[width=0.48\textwidth]{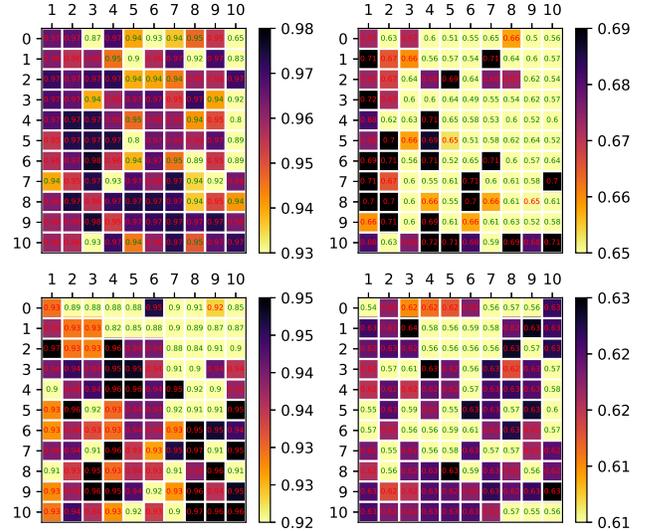}
    \caption{Heat map of the fusion model detection rate AUC (\%).}
    \label{fig:fusion}
    \vspace{-0.4cm}
\end{figure}
\subsection{Influence of Fusion Solutions}
We investigate the impact of view-level and feature-level fusion mechanisms on detection performance to respond the \textbf{RQ2}.
\subsubsection{View-level Fusion}As shown in Fig. \ref{fig:fusion}, the first and second row use Pubmed and Amazon Photo datasets respectively with two anomaly types (minimal class and synthetic method from left to right). Both of them calculate the detection rate on the hybrid model of GCN and GAT, the $x$ and $y$ axes represent their corresponding fusion weights. Darker colors, which corresponds to a higher detection rate, can be observed in the interior and on the diagonal of the heat map. This means that fusion between GCN and GAT methods does increase by 2 to 3 percents on the detection rate, but how to assign the optimal weights remains a problem. In this paper, we allow the neural network to learn how to assign the weights between different methods or views. However, the learned weights are still locally optimal. The following feature-level fusion can be considered as an alternative resolution for finding the optimal parameters. Still, it is worthwhile to explore more deeply on how to approximate the optimal weights.
\begin{figure}[t]
    \centering
    \includegraphics[width=0.48\textwidth]{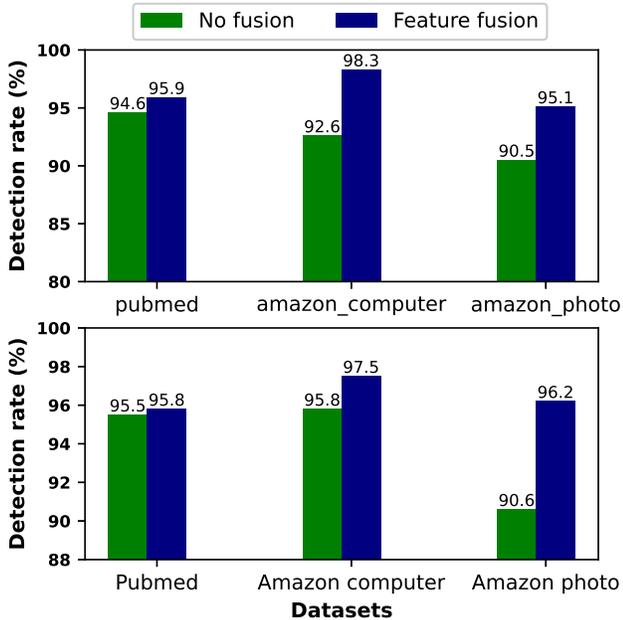}
    \caption{Comparison with no fusion and feature fusion on detection rate AUC (\%). The top one uses the combination of bwgnn, gin, while the bottom applies the fusion of bwgnn, gin and gat methods. Both of them verify on the minimal class anomalies of Pubmed, Amazon Computer and Amazon Photo.}
    \label{fig:feature_level_experiment}
    \vspace{-0.6cm}
\end{figure}
\subsubsection{Feature-level Fusion}
To explore the impact of feature-level fusion on performance, we compare unfused as well as feature-level fusion approaches as shown in Fig. \ref{fig:feature_level_experiment}. Experiments reveal that the method with feature-level fusion consistently increases by 2 to 3 on AUC percentage points. Even, the improvement is nearly 5 on the Amazon Photo. For the reliability of the results, we make experiments on the combination of bwgnn, gin and bwgnn, gin, gat method, which correspond to the top and bottom bar charts respectively. As explained previously, feature-level fusion works because it takes full advantage of the complementary information, while avoiding the impact of redundant information on detection performance.
% 最终的线性分类器，对于生成异常使用linear好
% 最终的线性分类器，对于最小异常使用GCNlinear比较好
% 小tricks，用简单的对比图体现。再简单说明一下
% \subsection{Influence of Classifier}
\begin{figure}[!t]
    \centering
    \includegraphics[width=0.48\textwidth]{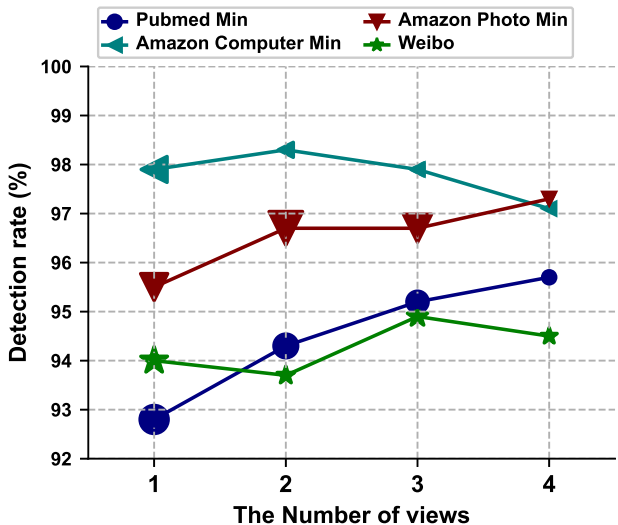}
    \caption{Line chart of the number of fusion methods on detection rate AUC (\%).}
    \label{fig:nums_views_methods}
    \vspace{-0.5cm}
\end{figure}
\subsection{Influence of The View Numbers}
To address the \textbf{RQ3}, we explore the impact of the number of fused views on detection performance. Since we are provided with the number of fused views instead of the specific fused views, it is tough to obtain an accurate AUC. Regarding this, we maintain a method pool containing all available methods, such as \{gin, gcn, gat, bwgnn\}. To get the detection AUC of 2 views, we randomly pick two methods from the method pool. The final AUC is the mean AUC over five times sampling. As shown in Fig. \ref{fig:nums_views_methods}, we conducted experiments on four datasets, where Pubmed Min refers to the minimal class anomalies of Pubmed. The trend of the line graph indicates that the detection AUC gradually increases as the number of fusion views rises. However, it is not the case that the more views are fused, the better the performance will be, one such case is minimal class anomalies of Amazon Computer. After 2 views, the detection AUC is gradually decreasing. It is noticeable that the size of each marker, which is used to express the standard deviation of the AUC, is varied. The increment in the number of fused views leads to a smaller marker, which means the algorithm is more stable.
\begin{table*}[!t]
\renewcommand{\arraystretch}{1.6}
\caption{Comparison with the shallow learning and GNN-based methods. The detecting results AUC$\pm$STD (Validation best results) (\%) over five seeds, among minimal class, synthetic and organic anomalies.}
\label{table.compare_with_sota}
\centering
\setlength{\tabcolsep}{1mm}{
\begin{tabularx}{\textwidth}{p{1.6cm}|cccccccc}
\Xhline{2\arrayrulewidth}
\multirow{2}{*}{\diagbox[]{Alg.}{Data.}}& \multicolumn{2}{c}{Pumbed}& \multicolumn{2}{c}{Amazon Computer}& \multicolumn{2}{c}{Amazon Photo}& \multirow{2}{2em}{Weibo} & \multirow{2}{2em}{Books}\\
& Min. & Syn.& Min. & Syn.& Min. & Syn.& & \\
\hline
IForest\cite{liu2008isolation_tree}&50.9$\pm$2.3(50.6)&	38.7$\pm$1.8(40.1)&48.1$\pm$0.8(48.3)&51.4$\pm$2.1(54.0)&	53.1$\pm$2.3(51.7)&51.9$\pm$2.3(54.0)&47.1$\pm$2.5(46.6)&	60.0$\pm$2.1(62.8)\\
LOF\cite{breunig2000lof}&51.8$\pm$0.0(52.1)&65.4$\pm$0.0(64.4)&	53.7$\pm$0.0(47.7)&	59.4$\pm$0.0(59.9)&68.8$\pm$0.0(67.0)&	65.7$\pm$0.0(57.7)&	61.5$\pm$0.0(59.9)&45.0$\pm$0.0(43.3)\\
HBOS\cite{goldstein2012hbos}&48.9$\pm$0.0(48.6)&34.3$\pm$0.0(30.8)&	50.0$\pm$0.0(42.6)&	47.1$\pm$0.0(47.1)&52.2$\pm$0.0(56.5)&	42.2$\pm$0.0(45.3)&28.0$\pm$0.0(33.8)&70.7$\pm$0.0(60.5)\\
OCSVM\cite{scholkopf2001oc_svm}&57.1$\pm$0.0(56.8)&	75.9$\pm$0.0(74.1)&	46.8$\pm$0.0(47.5)&	75.2$\pm$0.0(74.1)&51.5$\pm$0.0(50.6)&	75.9$\pm$0.0(76.0)&	80.7$\pm$0.0(83.5)&14.8$\pm$0.0(27.3)\\
\hline
GIN\cite{GCN}&91.3$\pm$0.1(91.6)&76.3$\pm$0.2(75.5)&	95.6$\pm$0.5(95.5)&50.5$\pm$8.9(60.1)&71.0$\pm$20.9(86.8)&	53.0$\pm$10.2(57.8)&94.5$\pm$0.5(95.1)&	50.0$\pm$0.0(52.1)\\
GAT\cite{GAT}&92.1$\pm$0.1(91.9)&81.7$\pm$0.9(82.9)&	98.9$\pm$0.0(99.5)&68.5$\pm$0.4(68.1)&	94.9$\pm$0.9(97.2)&	68.4$\pm$0.9(68.7)&89.4$\pm$2.6(90.0)&	50.0$\pm$0.0(56.2)\\
GCN\cite{GCN}&91.5$\pm$0.1(92.3)&80.6$\pm$0.2(80.6)&	99.2$\pm$0.0(98.5)&69.2$\pm$0.2(70.4)&	97.2$\pm$0.1(95.2)&	66.4$\pm$0.3(66.7)&97.4$\pm$0.2(98.2)&	50.0$\pm$0.0(50.0)\\
BWGNN\cite{BWGNN}&95.3$\pm$0.0(95.1)&	83.4$\pm$0.2(83.1)&\textbf{99.6$\pm$0.0(99.0)}&\textbf{74.7$\pm$0.4(77.8)}&	97.1$\pm$0.5(96.8)&	68.2$\pm$0.4(76.3)&	93.2$\pm$0.3(97.3)&	56.5$\pm$6.7(77.3)\\
\hline
Mul-GAD (Ours)&	\makecell[c]{\\\textbf{95.6$\pm$0.0(94.6)}}&	\makecell[c]{\\\textbf{91.3$\pm$0.6(88.9)}}&\makecell[c]{\\99.0$\pm$0.0(99.2)}&	\makecell[c]{\\74.3$\pm$2.2(75.4)}&\makecell[c]{\\\textbf{98.7$\pm$0.1(97.5)}}&\makecell[c]{\\\textbf{76.5$\pm$0.9(77.6)}} &\makecell[c]{\\\textbf{97.8$\pm$0.1(97.4)}}& \makecell[c]{\\\textbf{72.0$\pm$11.6(47.9)}}\\
\Xhline{2\arrayrulewidth}
\end{tabularx}}
\vspace{-0.3cm}
\end{table*}
\subsection{Comparison with existing methods}
We compare our method with two categories of baselines, shallow learning and GNN-based methods. To answer the \textbf{RQ4}, we conduct a series of experiments on five datasets, including minimal class, synthetic and organic anomalies, over five seeds. As shown in Table. \ref{table.compare_with_sota}, AUC is obtained through the test set and the content in the bracket refers to the best AUC for the validation set, which spot the potential of the algorithm. It is not the case that the more views are fused, the better the model performance will be. Thus, we regard the best results in the combination [gat,gin], [bwgnn, gat], [bwgnn, gin] and [bwgnn, gat, gin] as the final AUC for our method. Experimental results show that our Mul-GAD method outperforms the other detection methods on human-constructed anomalous datasets (Pumbed, Amazon Photo). To make the results more convincing, we use the organic anomalous datasets (Weibo and Books) for inspection. Although our detection rate is not the highest on Amazon Computer, the gap is not large at least. Moreover, we promote the detection rates by a large margin on the synthetic anomalies of Pubmed ,Amazon Photo and the organic anomalous dataset Books. In summary, our Mul-GAD method approach outperforms the state-of-the-art not only on detection performance, but also in terms of generalization across the majority of datasets compared to the existing state-of-the-art methods.
\section{Conclusion}
To address the challenge that existing methods are poorly generalized on most datasets, we propose a multi-view fusion algorithm for graph anomaly detection (Mul-GAD). Our fusion strategies include view-level and feature-level. View-level fusion learns the contribution of each view to the detection performance, while feature-level fusion is utilized to model intuition that complementary information shall be made full use and redundant information ought to be neglected. Moreover, we summarize the objective functions for the existing graph anomaly detection methods and further analyze the impact of different objective functions. For the integrity, the number of fused views is investigated. Exploiting these experimental evidences, we propose the Mul-GAD equipped with fusion strategies and label-oriented objective function. A theoretical justification of the fusion strategy is provided via Venn diagram for rigorous. Although our method outperforms other state-of-the-art anomaly detection methods in most scenarios, the detectors are still faced with a more complex and unknown situations. Extending our method to more realistic settings (e.g., heterogeneous or dynamic graph) is crucial. Exploring how to approximate the optimal parameters for view-level fusion is also a worthwhile research subject.
\section*{Acknowledgments}
This work was supported by the National Key Research and Development Program of China (No.2021YFB2700600) and the National Natural Science Foundation of China Enterprise Innovation and Development Joint Fund (No.U19B2044).

\section*{Conflicts of Interest}
The authors declare that there are no conflicts of interest regarding the publication of this paper.

\bibliographystyle{IEEEtran}
\bibliography{ref}
\end{document}